\documentclass[acmsmall]{acmart}
\acmJournal{TOMM}

\usepackage{amsmath,amsfonts}
\usepackage{cleveref}
\usepackage{algorithmic}
\usepackage{algorithm}
\usepackage{booktabs}     
\usepackage{graphicx}     
\usepackage{pifont}       
\usepackage{caption}      
\usepackage{subfigure}
\usepackage{bbding}
\usepackage{pifont}
\usepackage[table]{xcolor}

\AtBeginDocument{%
  }

\setcopyright{acmlicensed}
\copyrightyear{2018}
\acmYear{2018}
\acmDOI{XXXXXXX.XXXXXXX}

\acmJournal{JACM}
\acmVolume{37}
\acmNumber{4}
\acmArticle{111}
\acmMonth{8}

\begin{document}

\title{Rethinking the Learning Paradigm for Facial
Expression Recognition}

\author{Weijie Wang}
\email{weijie.wang.3777@gmail.com}
\orcid{1234-5678-9012}
\affiliation{%
  \institution{University of Trento \& Fondazione Bruno Kessler}
  \city{Trento}
  \state{Trento}
  \country{Italy}
}

\author{Bo Li}
\affiliation{%
  \institution{vivo Mobile Communication}
  \city{Shanghai}
  \country{China}}
\email{njumagiclibo@gmail.com}

\author{Nicu Sebe}
\affiliation{%
  \institution{University of Trento}
 \city{Trento}
 \state{Trento}
  \country{Italy}
}
\email{niculae.sebe@unitn.it}

\author{Bruno Lepri}
\affiliation{%
 \institution{Fondazione Bruno Kessler}
 \city{Trento}
 \state{Trento}
 \country{Italy}
}
 \email{lepri@fbk.eu}

\renewcommand{\shortauthors}{Trovato et al.}

\begin{abstract}
  Due to the subjective crowdsourcing annotations and the inherent inter-class similarity of facial expressions, real-world Facial Expression Recognition (FER) datasets usually exhibit ambiguous annotations. 
  To simplify the learning paradigm, most existing methods convert ambiguous annotation results into precise one-hot annotations and train FER models in an end-to-end supervised manner. In this paper, we rethink the existing training paradigm and propose that it is better to use weakly supervised strategies to train FER models with original ambiguous annotations. Specifically, we model FER as a Partial Label Learning (PLL) problem, which allows each training example to be labeled with an ambiguous candidate set. To better utilize representation learning to boost label disambiguation in PLL, we propose to use the Masked Image Modeling (MIM) strategy to learn the feature representation of facial expressions in a self-supervised manner. Then, these feature representations are used to query the probabilities of ambiguous label candidates, which are used as the ground truth in the PLL label disambiguation process. Extensive experiments on RAF-DB, FERPlus, and AffectNet demonstrate that the PLL learning paradigm substantially outperforms the current fully supervised state-of-the-art approaches in FER. 
\end{abstract}

\ccsdesc[500]{Computing methodologies~Computer vision}

\keywords{Facial expression recognition, Partial label learning, Deep learning}

\received{20 February 2007}
\received[revised]{12 March 2009}
\received[accepted]{5 June 2009}

\maketitle

\newcommand{\wwj}[1]{{\color{red}#1}}
\newcommand{\zz}[1]{{\color{red}#1}}

\section{Introduction}
Aiming at analyzing and understanding human emotions, Facial Expression Recognition (FER) is an active field of research in computer vision with practical implications~\cite{wang2023turn, wang2024uvmap, wang2023zero, wang2025fully, li2025freeinsert, nie2024t2td} in social robotics, automatic driving, etc.
\begin{figure}[!htb]    
\centering
\includegraphics[scale=0.3]{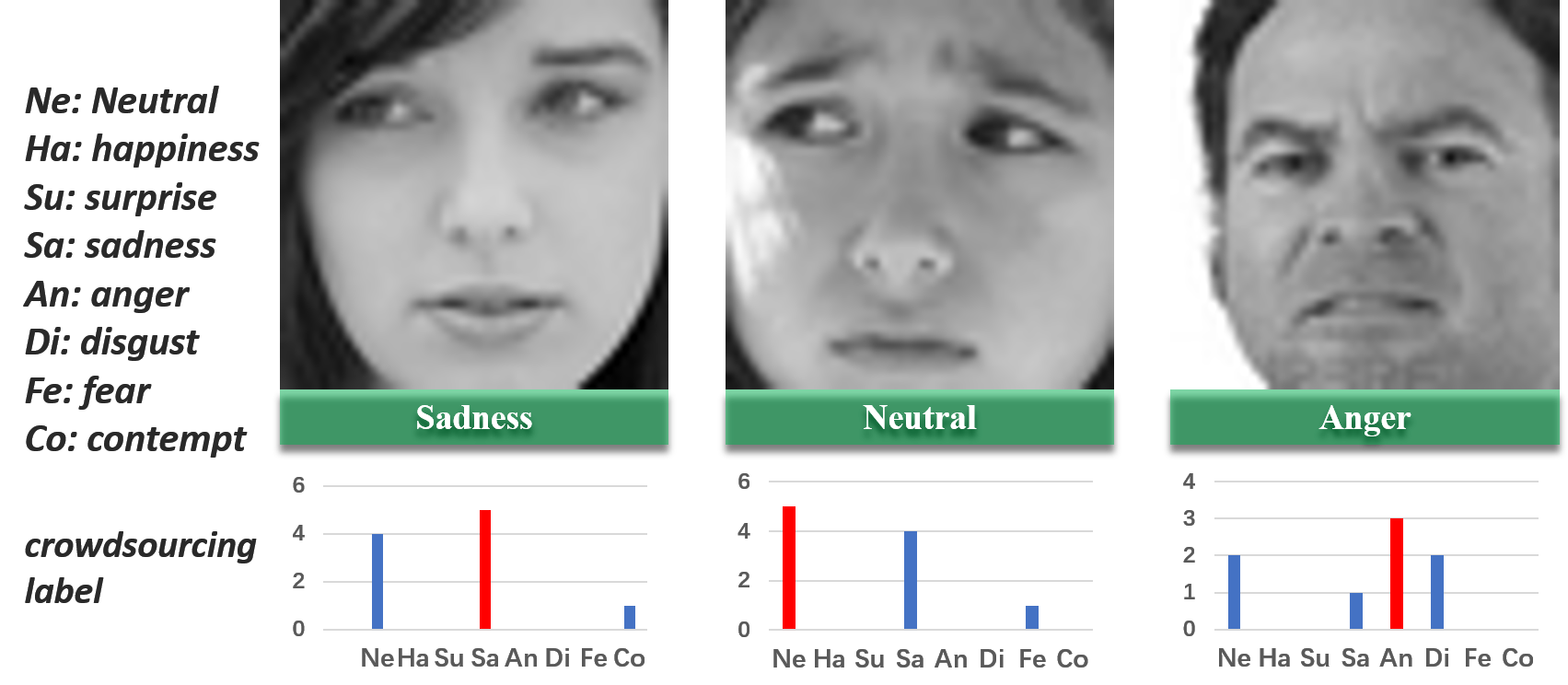}
\caption{
Random image samples from the FERPlus dataset with potentially noisy labels (colored in red) introduced by simple, voting-based label conversion~\cite{barsoum2016training} from crowdsourcing results.
}
\label{fig1}
\end{figure}
In recent years, FER has achieved impressive performance on laboratory-controlled datasets, such as CK+~\cite{lucey2010extended}, JAFFE~\cite{lyons1998coding} and RaFD~\cite{langner2010presentation}, which are collected under ideal conditions and annotated by experts. 
Recently, with the demands of real-world applications, large-scale FER datasets in unconstrained environments, such as FERPlus~\cite{barsoum2016training}, RAF-DB~\cite{li2017reliable} and AffectNet~\cite{DBLP:journals/corr/abs-1708-03985}, are required.

Annotating such large-scale datasets by experts is labor-intensive and challenging.
Consequently, crowdsourcing becomes increasingly popular.
Since the crowdsourced annotations are essentially a rating distribution for each sample, a typical way to leverage them is to convert them into a one-hot label by simple voting~\cite{barsoum2016training} or thresholding~\cite{li2017reliable} and then use it for a fully-supervised model training.
Since crowdsourcing usually yields subjective and ambiguous annotations, the above label conversion may yield ambiguous results as shown in Figure~\ref{fig1}, and the subsequent model training inevitably leads to sub-optimal FER models.

Therefore, recent work has started exploring how to learn with noisy labels. 
A typical solution is to derive a label distribution from the one-hot label and rely on distribution learning~\cite{wang2020suppressing, she2021dive}. 
However, this is limited since a potentially correct label may be assigned a relatively low weight, constraining the possibility of overriding the original, noisy label.



To address the above limitation, in this paper, we rethink the existing training paradigm of FER and propose to use weakly-supervised rather than the fully-supervised training.
Concretely, we model FER training as a \textit{Partial Label Learning (PLL)} problem~\cite{cour2011learning}, where a set of candidate labels is derived from the one-hot ground truth label.
This new paradigm avoids relying solely on a single label that is potentially noisy.
An important design in our PLL method is to adopt Masked Image Modeling (MIM)~\cite{he2022masked} for model pre-training.
This is in line with the consensus in PLL research that a higher-quality feature representation leads to a better PLL method~\cite{DBLP:journals/corr/abs-2201-08984,DBLP:conf/nips/LiuD12,zhang2016partial,lyu2019gm}.
The use of the Vision Transformer (ViT)~\cite{dosovitskiy2020image} in MIM particularly helps learn the local facial action units and global facial structures in various expressions~\cite{he2022masked}.
%

%
%

Moreover, in order to address the high inter-class similarity, we propose to treat the label disambiguation process as querying the confidence of the correlation between each category label and features.
Specifically, inspired by DETR~\cite{DBLP:conf/eccv/CarionMSUKZ20}, we propose a decoder-based label disambiguation module.
We leverage learnable label embeddings as queries to obtain confidence in the correlation between features and the corresponding label via the cross-attention module in the Transformer decoder.
Finally, we integrate the encoder of the ViT backbone, and the Transformer decoder-based label disambiguation module to build a fully transformer-based facial expression recognizer in the partial label learning paradigm.

Our contributions are summarized as follows:

\begin{itemize}
    \item  We rethink the existing training paradigm of FER and propose to use weakly supervised strategies to train FER models with original ambiguous annotation. To the best of our knowledge, it is the first work that addresses the annotation ambiguity in FER with the Partial Label Learning (PLL) paradigm.
    \item We build a fully transformer-based facial expression recognizer, in which we explore the benefits of the facial expression representation based on MIM pre-train and the Transformer decoder-based label confidence queries for label disambiguation.
    \item Our method is extensively evaluated on large-scale real-world FER datasets. Experimental results show that our method consistently outperforms state-of-the-art FER methods. This convincingly shows the great potential of the Partial Label Learning (PLL) paradigm for real-world FER.
\end{itemize}

\section{Related Work}
\subsection{Facial Expression Recognition}
Facial Expression Recognition (FER)~\cite{wang2020suppressing,ng2003sift,DBLP:journals/corr/abs-1708-03985,albanie2018emotion} aims to help computers understand human behavior and interact with humans. Early FER approaches mainly rely on hand-crafted texture or geometric descriptors, which perform well on controlled in-the-lab datasets. With the availability of large-scale in-the-wild datasets annotated via crowdsourcing, recent research has shifted to deep learning–based FER. However, such real-world datasets often contain ambiguous annotations due to subjective perception and the inherent similarity among expressions. Prior methods~\cite{DBLP:conf/mm/WangWL19,DBLP:conf/cvpr/YangCY18,vo2020pyramid,farzaneh2020discriminant,gera2021landmark} typically convert the annotation distributions into one-hot labels using majority voting~\cite{barsoum2016training} or thresholding~\cite{li2017reliable}, which inevitably discards ambiguity information and may introduce incorrect supervision, thereby limiting model performance. To mitigate this issue, SCN~\cite{wang2020suppressing} and RUL~\cite{zhang2021relative} attempt to relabel noisy samples by estimating confidence from clean samples, while DMUE~\cite{she2021dive} learns latent label distributions with sample-level uncertainty modeling. Nonetheless, these approaches fundamentally rely on ambiguous one-hot labels and often require estimating noise rates or adopting complex model designs, hindering their robustness in unconstrained scenarios.

More recent FER works further emphasize the necessity of modeling annotation uncertainty. 
{DE-FER}~\cite{10375788} leverages masked image modeling to obtain discriminative and robust representations under noisy labels, and 
{FSLDL}~\cite{shin2024noisy} explicitly learns face-specific label distributions to characterize annotation noise. 
Similarly, 
{MAE-Face}~\cite{ma2023unified} shows that masked autoencoding benefits affective representation learning, whereas 
{TA-FER}~\cite{10149390} proposes a transformer-augmented architecture with online label distillation to refine uncertain labels. 
Beyond noise robustness, open-set FER has also been explored; 
{OPFER}~\cite{zhang2024open} addresses ambiguous or unseen expressions arising in the wild, and 
{MCR}~\cite{tan2025mining} adopts label-free consistency regularization to handle noisy supervision without explicit labels. 
Despite their improvements, these methods ultimately reduce ambiguous annotations to pseudo one-hot targets, leaving the original label uncertainty underutilized.
Ada-CM~\cite{li2022towards} further explores weak supervision by incorporating unlabeled data through adaptive confidence margins. 

However, existing approaches still do not directly preserve or exploit the ambiguous annotation sets produced by crowdsourcing. This gap motivates revisiting the FER training paradigm and exploring methods that can naturally operate on ambiguous annotations without enforcing deterministic supervision.

\begin{figure*}[!htb]    
\centering
\includegraphics[scale=0.45]{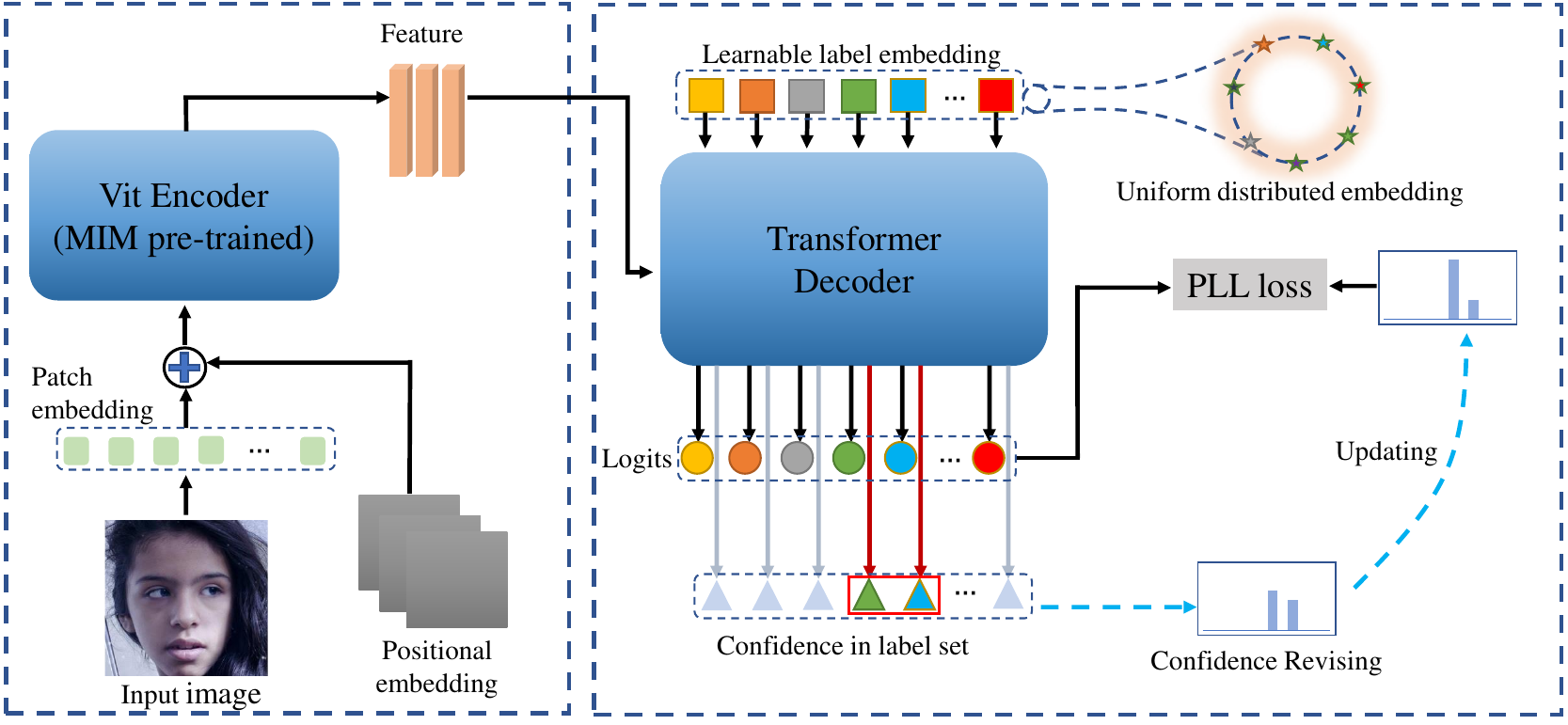}
\caption{
The overview of our framework. 
First, we use the MIM pre-trained ViT encoder as the backbone for representing the features of a 2D image. Second, we input the obtained feature from the left part to the transformer decoder with learnable label embedding. Third, we revise $(i-1)$-th confidence and update it to get $i$-th confidence. Finally, the loss is computed between logits and the $i$-th confidence.}\label{framework}
\end{figure*}

\subsection{Partial Label Learning}
Partial Label Learning (PLL) is a key weakly supervised learning method, where each object has multiple candidate labels, but only one of them is the ground-truth label. 
It is mainly divided into average-based and identification-based methods. The intuitive, average-based methods generally consider each label to be equally important during training and predict by averaging the output of all candidate labels~\cite{hullermeier2006learning,zhang2015solving}. Other studies~\cite{cour2011learning, zhang2016partial} use parametric models to maximize differences between the average scores of candidate labels and those of non-candidate labels. 
Identification-based methods normally maximize the output of the most likely candidate labels in order to disambiguate the ground-truth label, mainly by maximum margin criterion~\cite{yu2016maximum} or graph-based approaches~\cite{wang2021adaptive}. 
Deep learning advances lead to the emergence of network structures~\cite{wen2021leveraged,
feng2019partial}, rely on model output to disambiguate candidate label sets.
Some researchers claim~\cite{liu2012conditional,zhang2022semi,lyu2019gm, wu2022revisiting} that nearby data points in the feature space tend to share the same label, emphasizing the importance of a robust feature representation.
PICO~\cite{wang2022pico} reconciles PLL-based label disambiguation and feature representation through contrastive learning. However, elaborately designed data augmentation is essential due to its sensitivity to positive sample selection. 
Another drawback of contrastive learning in FER is that the high inter-class similarity of facial expressions makes it hard to obtain robust feature representations.

\subsection{Masked Image Modeling}
Inspired by Masked Language Modeling (MLM), Masked Image Modeling (MIM) becomes a popular self-supervised method. iGPT~\cite{chen2020generative} predicts unknown pixels by operating on known pixel sequences.
ViT~\cite{dosovitskiy2020image} predicts masked patch prediction by employing a self-supervised approach. 
Moreover, iBOT~\cite{zhou2021ibot} achieves excellent performance through siamese networks. 
However, these methods have the assumption of image semantic consistency. 
The ability of BEiT~\cite{bao2021beit} to predict discrete tokens depends on the pre-trained model VQVAE~\cite{van2017neural} that it relies on. 
As an extension of ViT, the main aim of SimMIM~\cite{xie2022simmim} is to predict pixels. Unlike the previously mentioned approaches, MAE~\cite{he2022masked} proposes an encoder-decoder architecture that uses the decoder for the MIM task. Differently, Maskfeat~\cite{wei2022masked} adopts hog descriptors as the prediction target rather than pixels. Currently, the potential benefits of MIM for enhancing PLL performance have not been well studied.

\section{Proposed Methods}
In this section, we introduce a fully transformer-based facial expression recognizer with a new partial label learning paradigm, which mainly consists of a process of robust learning representation learning and label disambiguation as shown in Figure \ref{framework}. Specifically, we explore the benefits of the facial expression representation based on MIM pre-train and the Transformer decoder-based label confidence queries for effective label disambiguation, on which we elaborate in the next sections.

\subsection{Problem Formulation}
Generally, given an in-the-wild FER dataset $\mathcal{D}(\mathcal{X},\mathcal{Y})$, each sample $x_i$ is assigned to one of the $K$ categories, denoted as $y_{i} \in \left \{y_{1}, y_{2},..., y_{k}\right \}$ which represents the deterministic category. This simplified form for labels is consistent with the supervised learning setting. However, as shown in Figure \ref{fig1}, most of the FER data is annotated via crowdsourcing, which can introduce labeling noise. Therefore, it is not advisable for most methods to oversimplify the FER task as a \textit{supervised learning} (SL) problem and apply one-hot labels to all annotated data while ignoring the ambiguity inherent in the label level.  

\textit{Partial-label learning} (PLL) is a multi-class classification problem.
Unlike traditional ones, PLL aims to predict a true label of the input using a mapping of the classification function. 
However, PLL can tolerate more uncertainty from the label space, and gradually eliminate it in a weakly supervised way during the training.
We argue that modeling the FER task as partial label learning is more appropriate. 
PLL assumes that each input $x_i$ must have a true label $y_i$ in the corresponding candidate label set $Y_i$ as the Eq.(\ref{eq1}):
\begin{equation}
    p(y_i \in Y_i |x_i,Y_i) = 1, \forall (x_i,y_i) \in \mathcal{X} \times \mathcal{Y}, \forall Y_i \in \mathcal{S},  
    \label{eq1}
\end{equation}
where $\mathcal{S}$ is the collection of all candidate label sets.
In addition, different from SL setup, the definition of each sample $x_i$ in partially labeled dataset $\tilde{\mathcal{D}} $ is denoted by $ (x_i, Y_i)$, where $Y_i$ is a set of candidate label corresponding to $x_i$. 

Thus, the aim of partial label learning (PLL) is to recognize the (unseen) true label $y_i$ of each sample $x_i$ from $Y_{i} = \left \{y_{1}, y_{2},..., y_{k}\right \}$ by training a classifier $\zeta :\mathcal{X} \to \mathcal{C}^k $, which utilize to compute the probability of each class in the candidate label set. We construct a candidate label set $Y_i$ for the sample $x_i$ in the FER dataset, where $Y_i$ is a set of binary codes. e.g. $y_{k}$ = 1 if the crowdsourcing result contains this category, else $y_k=0$. 
Before training, we normalized each label set $Y_i$ to obtain a confidence vector $C_i= \left \{c_{1}, c_{2},..., c_{k}\right \}$, which represents the probability value of the sample $x_i$ in the different categories, it sums to 1.
%
In the training stage, we train a classifier $\zeta :\mathcal{X} \to \mathcal{C}^k $ to update the confidence vector $c_i$. Ideally, the $c_i$ will concentrate most of the probability density on the true label $y_i$ of the sample $x_i$. We use the categorical cross-entropy loss to constrain it as the Eq.(\ref{eq2}) following:
\begin{equation}
\begin{aligned}
\hspace{-2mm}
\mathcal{L}_{pll} (x_i ,C_i) = \sum_{j=1}^{k} - C_i log( \zeta ^{i} (x_i)), \;  s.t. \; \sum_{j=1}^{k} c_{i,j} =1,
\label{eq2}
\end{aligned}
\end{equation}
where $i$ denotes the index of training sample $x_i$ and $j$ denotes the index of confidence vector $C_i$. And $\zeta^i(x_i)$ is the soft-max layer of the model with the input $x_i$. In the next section, we continue to describe the feature representation part of our proposed method.

\subsection{PLL with Pre-trained Feature Representation}

\begin{figure}[!htb]    
\centering
\includegraphics[scale=0.31]{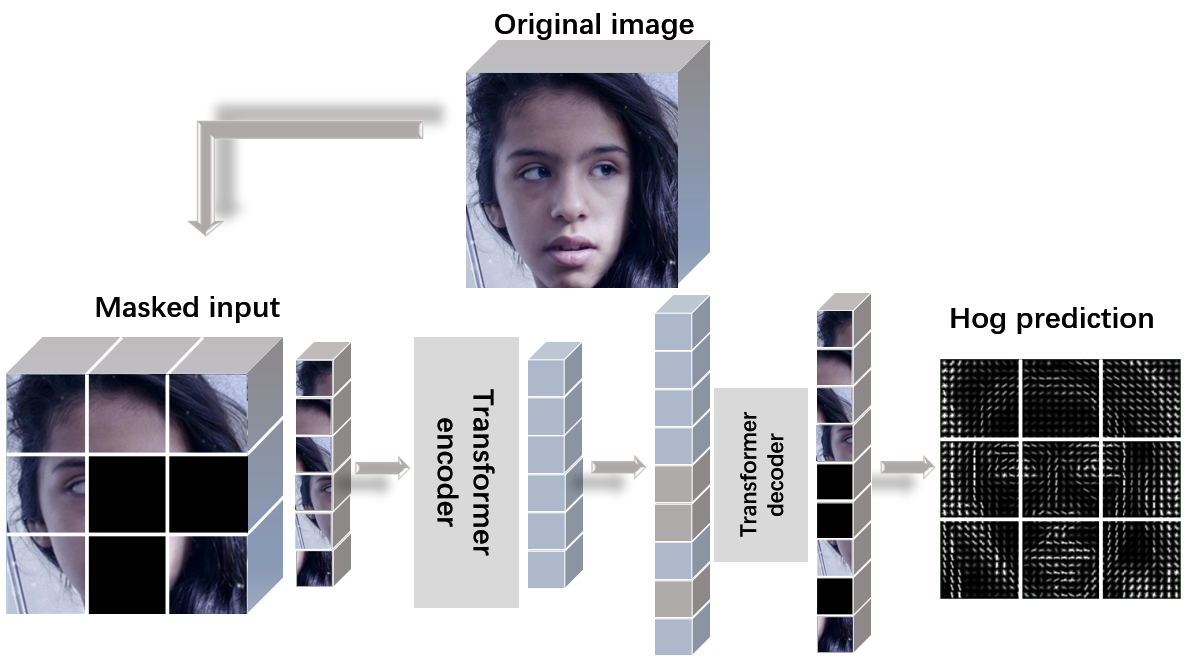}
\caption{We adopt Masked Image Modeling (MIM) for pre-training, which involves predicting the hog descriptor (HOG) with the randomly masked input. After obtaining the pre-trained model, we fine-tune it for the FER task within our framework.}\label{pretrain}
\end{figure}

In this work, we use MIM pre-train to obtain a better initial feature representation and based on that, use PLL to do label disambiguation. 
%
An important assumption of PLL is that samples close to each other in the sample space are more likely to share an identical label in the label space, which indicates that feature robustness is important for label disambiguation. 
However, as shown in Figure~\ref{fig1}, the uncertainty of the FER data leads to ambiguity at the label level, which hinders the feature representation.
Based on these observations, as shown in Figure~\ref{pretrain}, we use the MIM pre-training method to optimize feature representation.
Furthermore, since the network is easily misled by the incorrect labels, guiding feature learning by labels alone is difficult for the FER task, which is one of the bottlenecks of most current fully-supervised learning methods. 
Therefore, we argue that using a self-supervised pre-training manner can help in obtaining a better feature representation by avoiding being misled by incorrect labels.

Inspired by previous work~\cite{he2022masked,wei2022masked}, we adopt Masked Image Modeling (MIM) to pre-train facial expression representation by predicting hod descriptor. 
Specifically, in the encoder part, we adopt the same structure with ViT~\cite{dosovitskiy2020image} to pre-train, which is only for visible, unmasked patches. This is because the MIM pre-training manner forces the ViT to learn the local facial action units and global facial structures in various expressions. 
In FER, focusing on the variations in the action units of the facial expression is more meaningful rather than the information embedded in the pixel level. 
Thus, in the decoder part, we adopt hog descriptors as the prediction target of the decoder rather than pixel values \cite{he2022masked}.
In addition, predicting pixels is easy to influence by redundant information (high-frequency details, lighting, and contrast variations), which is less important in the FER task. 
In contrast, hog descriptor~\cite{dalal2005histograms} is dedicated to describing the distribution of edge directions within a local subregion. Finally, the output of the decoder can be reshaped to a reconstructed hog prediction of the image. 
In the training stage, we only keep the encoder for feature extraction and discard the decoder part.


\subsection{Label Disambiguation 
}
The label disambiguation with the transformer decoder mainly includes the label embedding regularization module and confidence revision module.
An input image $x$ is fed into the pre-trained encoder to get visual token $\mathcal{F}^{HW\times d}$, where $HW$ denotes the number of visual tokens and $d$denotes the feature dimension of the hidden state. We initialize the query embedding $\mathcal{Q}^{k \times d}$, where $k$ denotes the number of classes. In the training stage, the transformer decoder updates $\mathcal{Q}^{k \times d}$ according to visual tokens, which reflects the correlation of category-related features. Inspired by previous work~\cite{carion2020end,liu2021query2label}, we link the category-related features to the learnable query embedding $\mathcal{Q}^{k \times d}$ and predict the confidence sector $C_i$ of the sample by their correspondence. Specifically, $C_i$ is obtained by inputting the logits from the last layer of the transformer decoder to the sigmoid function. However, differently, (1) Our method is designed for PLL and focuses on disambiguation by query to finally find the unique correct label with the highest confidence for an instance. (2) Besides querying the classification logits, we also query the label confidence. (3) 
To mitigate the problem of imbalance distribution in the FER dataset, we design a special regularization for label embedding.

\subsubsection{Label Embedding Regularization}
It is worth noting that in addition to the ambiguity annotation these FER datasets also prevalently exhibit imbalanced distribution. In the training stage, it makes the model more biased towards majority classes, resulting in an uneven distribution of the feature space, and degrading model performance by classifying minority classes as head classes. 
In our work, we draw inspiration from ~\cite{wang2020understanding} and aim to achieve a uniform distribution of the query embedding over a hypersphere to mitigate the negative effects of class imbalance. Specifically, we compute a uniform loss for query embedding to pull the distance between different classes on the hypersphere. Specifically, we compute the optimal positions for different classes of query embedding $\left\{t_{i}\right\}_{i=1}^{K}$ before training and optimize it using the following Eq.(\ref{eq3}):

\begin{equation}
\mathcal{L}_{uniform}\left(\left\{t_{i}\right\}_{i=1}^{K}\right)=\frac{1}{K} \sum_{i=1}^{K} \log \sum_{j=1}^{K} e^{t_{i}^{T} \cdot t_{j} / \tau},
\label{eq3}
\end{equation}
where $\tau > 0$ is a temperature hyperparameter, $K$ denotes the number of classes in FER dataset. 
We explore the impact of $K$ in the \cref{app:K}.

\subsubsection{Revision Confidence}
As previously mentioned, the robust representation in the feature space enhances PLL's ability to disambiguate in the label space. For samples with ambiguous labels in the dataset, we initially set the pseudo targets for candidate label set $Y$ with a uniform distribution, represented by $C_i = \frac{1}{|Y_i|} Y_i $. During the update process, we continuously revise the confidence score $C_i$ for each batch using the following Eq.(\ref{eq4}):

\begin{equation}
\binom{id_{1}:v_{1}}{id_2:v_2} = Top_{k}[\varphi * C_{i-1}], k=2 ,
\label{eq4}
\end{equation}
where $\varphi=sigmoid(\sigma_{logits})$ is the response of sigmoid layer, and $\sigma _{logits}$ is the output of the transformer decoder. After that, we compute the $top2$ metrics of the $(C_{i-1} * \varphi )$ for updating confidence vector $C_i$ as the Eq.(\ref{eq5}):


\begin{equation}
C_i = 
\left\{ \begin{array}{lcr} 
C_{i-1}          \;\; \quad  \quad \quad \quad \quad \quad \Delta{v} \le  threshold \\
C_{i-1}[id_1] = 1 \quad \quad \quad    else
\end{array} \right. , \label{eq5}
\end{equation}
where $\Delta {v} = |v_{1} -  v_{2}| $, the threshold $\in$ [0,1). 
When $\Delta{v} > $ threshold, $C_i$ is a one-hot vector, where only the 1-th entry is 1.
We treat it as the level of uncertainty and represent it with the top2 rank label in the candidate label set. 
Algorithm \ref{Algorithm} is the pseudo-code of our method, which would converge to a stable status along the training iterations.

\begin{algorithm}[!t]
  \caption{Pseudo-code of our method} 
  \begin{algorithmic}
  \REQUIRE
      $Model\; f$: Pre-trained model $f$;
    \STATE
      $Epoch\; T_{max}$: Number of total epochs;
    \STATE
      $Iteration\; I_{max}$: Number of total iterations;
    \STATE
    $query\;embedding$: query $\mathcal{Q}$ ;  
    \STATE
      $Dataset \: \tilde{\mathcal{D}}$ : partially labeled training set $ \tilde{\mathcal{D}}=\left \{ (x_i, Y_i) \right \}^{n}_{i=1}$;
    \ENSURE
       \STATE
        $C_i = \frac{1}{|Y_i|} Y_i (i \in Y) $, initialize the confidence vector $C_i$;
        \FOR{$i = 1,2,..., T_{max}$}
      \STATE  \textbf{Shuffle} $ \tilde{\mathcal{D}}=\left \{ (x_i, Y_i) \right \}^{n}_{i=1}$ ;
      \FOR{$j = 1,2,..., I_{max}$}
      \STATE \textbf{fetch} mini-batch $ \tilde{\mathcal{D}}_j$ from $ \tilde{\mathcal{D}}$;
      \STATE \textbf{calculate} Uniform distribution for query by Eq.(\ref{eq3});
      \STATE \textbf{revise} confidence vector $C_i$ by Eq.(\ref{eq4}) and Eq.(\ref{eq5}); 
      \STATE \textbf{calculate} PLL loss by Eq.(\ref{eq2})
      \STATE \textbf{update} Network parameter of $f$ by Eq.(\ref{eq2});
      \ENDFOR
      \ENDFOR
  \end{algorithmic}
  \label{Algorithm}
\end{algorithm}

\section{Experiments}

In this section, we first describe the FER datasets and experiment implementation details. To evaluate our proposed method, we compare our method with baseline and other related methods including
\textbf{fully-supervised learning}: PSR~\cite{vo2020pyramid}, DDA~\cite{farzaneh2020discriminant}, SCN~\cite{wang2020suppressing}, SCAN~\cite{gera2021landmark}, DMUE~\cite{she2021dive}, RUL~\cite{zhang2021relative}, DACL~\cite{farzaneh2021facial}, MA-Net~\cite{zhao2021learning}, MVT~\cite{li2021mvt}, VTFF~\cite{ma2021facial}, EAC \cite{zhang2022learn}, TransFER~\cite{xue2021transfer}; 
\textbf{partial label learning}: RC~\cite{feng2020provably}, LW~\cite{wen2021leveraged}, PICO~\cite{wang2022pico}, CRPLL~\cite{wu2022revisiting}.

\subsection{Datasets}
\noindent \textbf{RAF-DB}~\cite{li2019reliable} contains around 30K diverse facial images with single or compound expressions labeled by 40 trained annotators via crowdsourcing. In our experiments, we use 12271 images for training and 3068 images for testing.

\noindent \textbf{FERPlus}~\cite{BarsoumICMI2016} is an extended dataset of the standard emotion FER dataset with 28709 training, 3589 validation, and 3589 test images. It provides a set of new labels for each image labeled by 10 human annotators. According to previous work, we just use the training part and test part.

\noindent \textbf{AffectNet}~\cite{DBLP:journals/corr/abs-1708-03985} is one of the most challenging FER datasets with manually labeled 440k images. As per prior research, we divide it into AffectNet-7 and AffectNet-8, with the latter having one more expression of contempt and 3667 training and 500 test images compared to the former.

\begin{table*}[t]
\centering
\begin{minipage}{0.48\linewidth}
\centering
\caption{\textbf{Pre-training setting, 
Part \uppercase\expandafter{\romannumeral1}.}}
\label{exp:table:pre_1}
\begin{tabular}{l|ll}
 configs & \textbf{MAE} & \textbf{MaskFeat} \\ 
\hline
Base Learning Rate & 1.5e-4 & 2.5e-5 \\
Batch\_Size & 1024(4096) & 1024(2048) \\ 
Warm Up & 40 & 40 \\
Weight\_Decay & 0.05 & 0.05 \\
Norm\_Pix\_Loss & True & False \\
Mask-Ratio & 85\%(75\%) & 85\%(40\%) \\
Epoch & 600 & 600 \\
Optimizer & AdamW & AdamW \\
Optimizer Momentum & 0.9,0.95 & 0.9,0.999 \\
Mask\_Patch & 16 & 16
\end{tabular}
\end{minipage}
\hfill
\begin{minipage}{0.48\linewidth}
\centering
\caption{\textbf{Pre-training setting, 
Part \uppercase\expandafter{\romannumeral2}.}}
\label{exp:table:pre_2}
\begin{tabular}{l|ll}
 configs & \textbf{MoCo} & \textbf{SimMIM} \\ 
\hline
Base Learning Rate & 0.03 & 1.5e-4(1e-4) \\
Batch\_Size & 256 & 1024(2048) \\ 
Warm Up & 0 & 40 \\
Weight\_Decay & 1e-4 & 0.05 \\
Norm\_Pix\_Loss & False & False \\
Mask-Ratio & 0 & 75\%(60\%) \\
Epoch & 200 & 800 \\
Optimizer & SGD & AdamW \\
Optimizer Momentum & 0.9 & 0.9,0.95 \\
Mask\_Patch & 0 & 16(32)
\end{tabular}
\end{minipage}
\end{table*}

\begin{table}
\centering
\label{exp:table:fine_1}
\caption{\textbf{Fine-tuning setting of Maskfeat on FER datasets.}}
\scalebox{0.9}{
\begin{tabular}{l|lll}
 configs & \textbf{RAF-DB} & \textbf{FERPlus} &\textbf{AffectNet-7/8}\\ 
\hline
Base Learning Rate & 1e-4  & 1e-4 & 1e-4 \\
Batch\_Size & 320 & 320 & 320 \\ 
Weight\_Decay &5e-2 & 5e-2 &5e-2\\
Mask-Ratio & 40\% 40\% & 40\% \\
Epoch & 100 & 100 & 25  \\
Optimizer &AdamW &  AdamW& AdamW \\
OptimizerMomentum &0.9,0.95 & 0.9,0.95&0.9,0.95 \\
\end{tabular}
}
\end{table}

\subsection{Baselines and Experiment Setup}
We adopt ViT-B/16~\cite{dosovitskiy2020image} pre-trained on Imagenet-1K as our baseline. For fairness, we keep the same parameters as our fine-tuning part in all settings, which will be described later. In terms of SL methods, we adopt their optimal settings to make fair comparisons. For data preprocessing, we keep all the images with the size of $224 \times 224$ and use MobileFacenet~\cite{chen2018mobilefacenets} to obtain aligned face regions for RAF-DB and FERPlus. For AffectNet, alignment was obtained via the landmarks provided in the data. 
In the pre-training part, we use four Tesla V100 GPUs to pre-train the ViT for 600 epochs with batch size 192 and learning rate 2e-4. We set the mask ratio of Maskfeat to 0.4. In fine-tuning part, we use only two Tesla V100 GPUs. We utilize the AdamW optimizer, setting the batch size to 320 and the initial learning rate to 1e-4 with the weight decay of 5e-2. The pre-training and fine-tuning protocol that we use is based on~\cite{wei2022masked}.  
In the PLL setup, for the FER dataset without crowdsourcing results, we construct the candidate set for it based on LW~\cite{wen2021leveraged}.

\subsubsection{Details for pre-training and fine-tuning}
\textbf{Pre-training and Fine-tuning.}
Table \ref{exp:table:pre_1} and Table \ref{exp:table:pre_2} summarizes the pre-training configurations of different pre-trained methods. 
~\Cref{exp:table:fine_1} shows the fine-tuning configurations on the FER dataset. 
We use three FER datasets to pre-train the encoder part and adopt 2 transformer blocks in the decoder part.

\subsubsection{Time/Memory Consumption of Proposed Method}
We trained 600 epochs using 4 46GB Tesla A40, requiring ~1.5 days for pre-training, which takes longer and relatively more memory resources. 
The related pre-trained checkpoint will be released to help with FER downstream tasks. 
As for fine-tuning, we do an ablation on RAF-DB with batch size=72/128/256, and it takes around 1.43/1.2/1.18 hours and around 11/16/30 GB of memory, which is similar to DMUE\cite{she2021dive})/MA-NET\cite{zhao2021learning}/EAC\cite{zhang2022learn}. However, we use similar resources to achieve better performance.

\subsection{Comparison with the SOTA FER Methods}
We conduct extensive comparisons with the SOTA FER methods on RAF-DB, FERPlus, AffectNet-7, and AffectNet-8, respectively. Specifically, we compared \textbf{ResNet18-based} method, which is pre-trained on face recognition dataset MS-Celeb-1M~\cite{guo2016ms}, such as SAN~\cite{wang2020suppressing}, DMUE~\cite{she2021dive}, RUL~\cite{zhang2021relative},
DACL~\cite{farzaneh2021facial},
MA-Net~\cite{zhao2021learning},
DDA~\cite{farzaneh2020discriminant}, EAC~\cite{zhang2022learn}.
\textbf{ResNet50-based}: SCAN~\cite{gera2021landmark},
{DR-FER~\cite{10375788}, and FSLDL~\cite{shin2024noisy}}. 
In addition, we also compare \textbf{Transformer-based} such as MVT~\cite{li2021mvt}, VTFF~\cite{ma2021facial}, TransFER~\cite{xue2021transfer},
{MAE-Face~\cite{ma2023unified}, and TA-FER~\cite{10149390}}. 
And only PSR~\cite{vo2020pyramid} using \textbf{VGG-16}. 
We list all the comparison results in Table \ref{table1}. 
We achieve the highest performance on all datasets as shown in Table \ref{table1}. 
From the results, our method outperforms all current SOTA methods for FER, including ResNet-based and Transformer-based methods. 
Specifically, we exceed the baseline~\cite{dosovitskiy2020image} by 3.69\%, 2.93\%, 2.5\%, 2.78\% accuracy on RAF-DB, FERPlus, AffectNet-7, AffectNet-8.

\begin{table*}[!ht]
  \centering
  \caption{Comparisons with the state-of-the-art methos on the RAF-DB, FERPlus, AffectNet-7 and AffectNet-8, respectively. 
  Accuracy (\%) is reported on the test dataset. $^*$ denotes our implementation. ``PR" denotes pre-train.}
    \setlength{\tabcolsep}{2mm}{\begin{tabular}{cccccc}
    \toprule
    \multicolumn{1}{c}{Method}& \multicolumn{1}{c}{Backbone} & RAF-DB & \multicolumn{1}{c}{FERPlus} & AffectNet-7 & AffectNet-8 \\
    \midrule
    PSR~\cite{vo2020pyramid}         & VGG-16 & 88.98 & 89.75 & 63.77 & 60.68 \\
    DDA~\cite{farzaneh2020discriminant} & ResNet-18 & 86.90  & -     & 62.34 & - \\
    SCN~\cite{wang2020suppressing}   & ResNet-18 & 87.03 & 88.01 & 63.40$^*$ & 60.23 \\
    SCAN~\cite{gera2021landmark}     & ResNet-50   & 89.02 & 89.42 & 65.14  & 61.73 \\
    FDRL~\cite{Ruan_2021_CVPR}       & ResNet-18 & 89.42 & - & -  & - \\    
    DMUE~\cite{she2021dive}   & ResNet-18 & 88.76 & 88.64 & - & 62.84 \\
    RUL~\cite{zhang2021relative}     & ResNet-18 & 88.98 & - & 61.43 & - \\
    DACL~\cite{farzaneh2021facial}   & ResNet-18 & 87.78 & 88.39$^*$ & 65.20  & 60.75$^*$ \\
    MA-Net~\cite{zhao2021learning}   & ResNet-18 & 88.40 & - & 64.53  & 60.96 \\
    
    MVT~\cite{li2021mvt}       & DeIT-S/16      & 88.62 & 89.22 & 64.57 & 61.40 \\
    VTFF~\cite{ma2021facial}   & ResNet-18+ViT-B/32 & 88.14 & 88.81 & 64.80 & 61.85 \\
    TransFER~\cite{xue2021transfer} & CNN+ViT & 90.91 &  89.76&   66.23  \\
    EAC~\cite{zhang2022learn} & ResNet-18 & 89.99 &89.64 & 65.32 & - \\
    DR-FER~\cite{10375788} & ResNet-50 & 90.61 & 91.91 & 67.54 & 63.60 \\
    FSLDL~\cite{shin2024noisy} & ResNet-50 & 90.97 & - & 65.60 & - \\
    TA-FER~\cite{10149390} & ViT-B/32 & 89.12 & 90.67 & 65.17 & -  \\
    \midrule
    MAE-Face~\cite{ma2023unified} & ViT-B/32  & - & - & 69.51 & 66.65\\
    \midrule
    RC~\cite{feng2020provably} & ResNet-18      & 88.92  & 88.53 & 64.87  & 61.63  \\
    LW~\cite{wen2021leveraged} & ResNet-18      & 88.70  & 88.10  & 64.53  & 61.46 \\
    PICO~\cite{wang2022pico}   & ResNet-18      & 88.53  & 87.93 & 64.35   & 61.38 \\
    CRPLL~\cite{wu2022revisiting} &ResNet-18 &88.47&88.01&64.32&61.37\\ 
    \midrule
    Baseline~\cite{dosovitskiy2020image} & ViT-B/16 & 88.36 & 88.18 & 64.61 & 61.34 \\
    ours w/o PR  & ViT-B/16 & 90.91  & 89.76 & 65.23 & 62.45 \\
    \rowcolor{gray!20}
    Ours & ViT-B/16 & \textbf{92.05} & \textbf{91.11} & \textbf{67.11} & \textbf{64.12} \\
    \bottomrule
    \end{tabular}}%
  \label{table1}%
\end{table*}%

\subsubsection{Comparison with SL methods}
Compared with the latest SL methods, we surpass EAC~\cite{zhang2022learn} by 2.06\% and 1.79\% on RAF-DB and AffectNet-7 respectively, PSR~\cite{vo2020pyramid} by 1.36\% on FERPlus, and DMUE~\cite{she2021dive} by 1.28\% on AffectNet-8. 
These experimental results show the effectiveness of our proposed method in the FER task and demonstrate that converting ambiguous annotation results into one-hot labels using fully-supervised learning is not ideal. 
Compared to PLL, these CNN-based SL methods are misled by the incorrect labels in the label space, which hinders the performance of the model. 
Furthermore, the Transformer-based SL method suffers from the same problems. 
Although MVT~\cite{li2021mvt} employs ViT as the backbone, it does not utilize unsupervised pre-training. Nevertheless, it remains susceptible to incorrect labels in the label space, resulting in potential impacts on feature learning quality, which is consistent with the hypothesis proposed in \cite{liu2021self}. The same problem exists with VTFF~\cite{ma2021facial} and TransFER~\cite{xue2021transfer}.  
Other SL-based approaches, including SCAN~\cite{gera2021landmark}, DR-FER~\cite{10375788}, and FSLDL~\cite{shin2024noisy}, also fall short of our method across all datasets. 
For example, our approach outperforms SCAN by 3.03\%, 1.69\%, and 1.97\% on RAF-DB, FERPlus, and AffectNet-8, respectively. 
Similarly, although DR-FER benefits from advanced representation learning and achieves strong performance (e.g., 90.61\% on RAF-DB and 91.91\% on FERPlus), our method still improves over it by 1.44\% and 0.80\%, respectively. 
FSLDL is also clearly surpassed by our method on RAF-DB (90.97\% vs. 92.05\%), demonstrating that even stronger SL pipelines cannot compensate for the information loss caused by one-hot label conversion.
Furthermore, although Transformer-based SL approaches such as TA-FER~\cite{10149390} achieve competitive results, they still underperform our model across all comparable datasets.  

\subsubsection{Comparison with SSL methods}
Notably, MAE-Face~\cite{ma2023unified} performs large-scale masked-autoencoding pre-training on millions of external facial images, which essentially places it at a \textbf{foundation-model scale} rather than within the standard FER training regime. 
This makes direct comparison with supervised FER methods inherently unfair, as its performance is largely attributed to the capacity and generalization benefits brought by such extensive pre-training.

\subsubsection{Comparison with PLL methods}
We add several representative PLL methods: RC~\cite{feng2020provably}, LW~\cite{wen2021leveraged},
PICO~\cite{wang2022pico},
CRPLL~\cite{wu2022revisiting}. 
For a fair comparison, we use their optimal experimental settings. The results in Table~\ref{table1} are our implementation based on their open-source codes.
The general PLL method for label disambiguation, such as RC and LW relies on the robustness of the features, and from the results, for the FER task, RC, LW, and CRPLL are misled by incorrect labels in the label space, which affects disambiguation due to the lack of robust facial expression feature representation.
In addition, it is worth noting the result of PICO demonstrates that contrastive-based partial label learning does not apply to the FER task. As mentioned before, the contrastive-based approach is sensitive to data augmentation. Unlike objects, facial expressions have high inter-class similarities, which consist of five fixed facial action units. 
Whereas typical data augmentation usually works on the lighting and contrast variation of the image, which only leads to extra redundant information in the FER task, sometimes, even performance degradation.
\subsection{Ablation Study}

\begin{table}[t]
  \centering
  \caption{
  Comparing Softmax and vanilla PLL with different pre-training manners on FER without disambiguation module. ``-PR'' indicates pre-training.}
  \setlength{\tabcolsep}{0.3mm}{
    \begin{tabular}{ccccc}
    \toprule
    Method  & RAF-DB & FERPlus   & AffectNet-7 \\
    \midrule
    ResNet-18 + Softmax & 86.54& 87.37 & 63.86 & \\ 
    ResNet-18 + PLL  & 88.67 &  88.40  & 64.49 \\
    Maskfeat-PR  + Softmax  &87.85 & 87.92 & 64.31 \\
    \rowcolor{gray!20}
    Maskfeat-PR + PLL  &\textbf{90.35} &
    \textbf{90.24}&\textbf{66.12} \\ 
    \bottomrule
    \end{tabular}}
  \label{exp:table:pll_effectiveness}
\end{table}

\subsubsection{Effectiveness of PLL Paradigm}
To verify the effectiveness of our PLL paradigm, we only use the ResNet-18 pre-trained on FER dataset MS-Celeb-1M~\cite{guo2016ms} and pre-trained Maskfeat as the backbone. we compare the traditional Softmax Loss to Vanilla PLL Loss only using Eq.(\ref{eq2}) without any other modules and report the results in Table \ref{exp:table:pll_effectiveness}.
(1) Compared with Softmax loss, PLL loss exceeds it across the board, improving the recognition accuracy by 2.13\% on RAF-DB, 1.03\% on FERPlus, and 0.63\% on AffectNet-7 in terms of ResNet-18.
We experimentally verified that a good performance can be achieved even if only using ResNet-18 with the vanilla PLL loss. 
The ``Maskfeat-PR'' setting yields a similar conclusion. 
(2) Table \ref{table1} reveals that most ResNet-based methods employ supervised learning to represent the label space distribution for samples, which is susceptible to the influence of incorrect labels in the label space. From the results, simplifying FER as a supervised learning task is not optimal.

\subsubsection{Feature-level Effects of Pre-trained SL, SSL}

\begin{table}[!t]
  \centering
  \caption{
  Different pre-training manners with vanilla PLL on FER without disambiguation module. ``-PR'' indicates pre-training.}
  \setlength{\tabcolsep}{0.4mm}{
    \begin{tabular}{ccccc}
    \toprule
    Method  & RAF-DB & FERPlus   & AffectNet-7 \\
    \midrule
    ResNet-18 + PLL  & 88.67 &  88.40  & 64.49 \\
    MoCo + PLL  &88.59 &88.14 & 63.98\\
    SimMIM-PR + PLL  & 89.60 & 89.78 & 65.34 \\
    MAE-PR + PLL  &89.99 &89.70 & 65.79 \\
    \rowcolor{gray!20}
    Maskfeat-PR + PLL  &\textbf{90.35} &
    \textbf{90.24}&\textbf{66.12} \\ 
    \bottomrule
    \end{tabular}}
  \label{table3}
\end{table}

\begin{figure}[!htb]    
\centering
\includegraphics[scale=0.4]{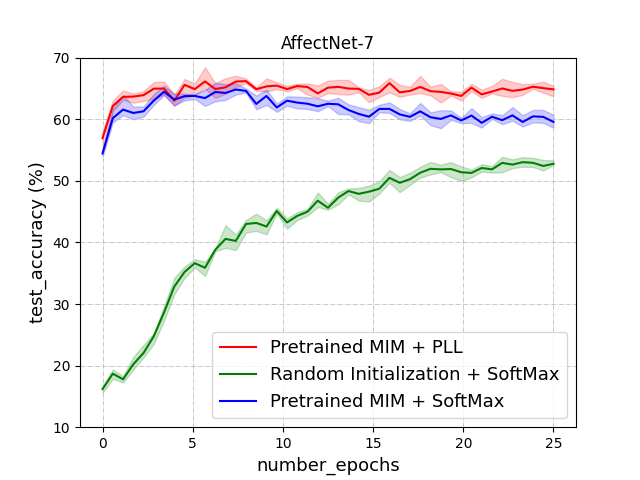}
\caption{Test accuracy during fine-tuning on AffectNet-7.}
\label{a7_acc}
\end{figure}

While vanilla PLL is effective, we demonstrate that robust features can significantly enhance the ability to disambiguate the PLL paradigm. We argue that self-supervised pre-training is the optimal method for obtaining robust features in the FER task. Thus, we also compare other SSL pre-training approaches (MoCo~\cite{he2020momentum},
SimMIM~\cite{xie2022simmim}, MAE~\cite{he2022masked}) as shown in Table \ref{table3}, and demonstrate that combining Maskfeat~\cite{wei2022masked} with PLL is a more effective self-supervised learning paradigm for FER. We note that the MoCo \cite{he2020momentum}, which is sensitive to data augmentation, is not very effective in the FER because the variations in contrast and illumination in the expression data have little effect on the facial structural units. As shown in Figure \ref{a7_acc}, we conduct a comparison between ``Pretrained MIM + PLL'', ``Pretrained MIM + SoftMax'' (which denotes a pre-trained encoder with a different loss function), and ``Random Initialization + SoftMax'' (which denotes random initialization of the encoder. (1) ``Random Initialization + SoftMax'' easily suffers from incorrect labels in the dataset during the whole fine-tuning, which further impedes the model from learning the high-quality features, thereby undermining the model's classification ability and decreasing its robustness. Compared to it, ``Pretrained MIM + SoftMax'' slows down overfitting to incorrect labels at the early stage, which is provably robust against the incorrect labels during fine-tuning. (2) At the initial stage of fine-tuning, ``Pretrained MIM + SoftMax'' also achieves superior performance. However, as training goes on, the interference of incorrect labels gradually increases, and the model begins to overfit these incorrect labels, ultimately resulting in performance degradation. Whereas, at the later stage of fine-tuning, ``Pretrained MIM + PLL'' resists the interference caused by incorrect labels, leading to robust performance in FER.

\begin{figure}[!htb]    
\centering
\includegraphics[scale=0.4]{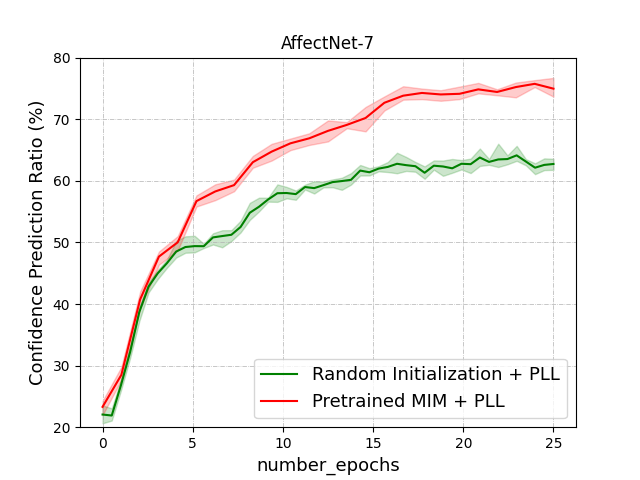}
\caption{The confidence correctness ratio of relabeled data on AffectNet-7.}\label{conf_acc}
\end{figure}

To quantitatively and intuitively observe and analyze the performance of PLL in terms of confidence, we artificially re-labeled and tested 2k samples in the AffectNet-7 (the ratio of incorrect labels is the highest in the FER dataset), and counted the confidence ratio of these samples, as shown in the Figure \ref{conf_acc}, during the fine-tuning, compared to ``Random initialization + PLL'', ``Pretrained MIM + PLL'' disambiguates these samples more effectively by combining the robust features obtained by SSL pre-training. They complement each other, but PLL plays a more important role in the disambiguation process.
We have attached some PLL disambiguation results as shown in ~\Cref{ferplusresult} on FERPlus. 
The experiment results demonstrate the effectiveness of the PLL paradigm. 
For RAF-DB, and AffectNet7, see \Cref{sup_raf}, and \Cref{sup_a7}, respectively.


\subsubsection{Evaluation of Decoder and key modules}
To evaluate the effectiveness of the transformer decoder and other key modules, we conduct an ablation study on RAF-DB and FERPlus as shown in Table \ref{table4}.
From the results, based on the PLL paradigm, the transformer decoder improve accuracy by 0.94\% on RAF-DB and 0.55\% on AffectNet7. The results show that it uses the cross-attention module to obtain the correlation of features and category labels via learnable label embedding, which in turn provides a more adequate foundation for the PLL disambiguation.

\begin{table}[t]
  \centering
  \caption{Evaluation of transformer decoder-based label disambiguation module with PLL. 
  Accuracy (\%) and gains over the baseline are reported. 
  ``Decoder'' denotes the transformer decoder, label embedding regularization is ``$\mathcal{L}_{uniform}$'', 
  and revision confidence is ``\emph{Re-conf}''.}
  \setlength{\tabcolsep}{0.7mm}{
  \begin{tabular}{cccc|cc|cc}
    \toprule
     & Decoder & $\mathcal{L}_{uniform}$ & \emph{Re-conf} 
     & RAF-DB & Gain & AffectNet-7 & Gain \\
    \midrule
    & \XSolidBrush & \XSolidBrush & \XSolidBrush 
      & 90.35 & -- & 66.12 & -- \\
    & \Checkmark & \XSolidBrush & \XSolidBrush 
      & 91.29 & \textcolor{red}{+0.94} & 66.67 & \textcolor{red}{+0.55} \\
    & \Checkmark & \Checkmark & \XSolidBrush  
      & 91.64 & \textcolor{red}{+1.29} & 66.90 & \textcolor{red}{+0.78} \\
    & \Checkmark & \XSolidBrush & \Checkmark 
      & 91.89 & \textcolor{red}{+1.54} & 66.93 & \textcolor{red}{+0.81} \\
    & \Checkmark & \Checkmark & \Checkmark 
      & \textbf{92.05} & \textcolor{red}{\textbf{+1.70}} 
      & \textbf{67.11} & \textcolor{red}{\textbf{+0.99}} \\
    \bottomrule
  \end{tabular}}
  \label{table4}
\end{table}

As for the label uniform embedding module, $\mathcal{L}_{uniform}$ alleviates the influence of imbalanced FER data classes by distributing query embedding uniformly over the hypersphere.
As for the revision confidence module, the revision confidence strategy can further facilitate the ability of PLL to disambiguate in the label space. This module enhances performance as an auxiliary module of disambiguation. 

\subsubsection{Ablation for Threshold of Revision Confidence}
\label{exp:aba:thre_revision_conf}
\begin{figure}[!htb]
    \centering
    \subfigure[The accuracy (\%) with different thresholds on RAF-DB and FERPlus.]{
        \includegraphics[scale=0.45]{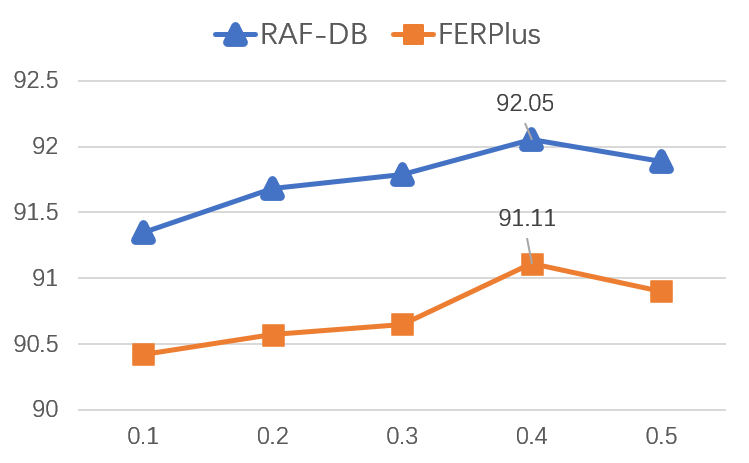}
        \label{threshold_rf}
    }
    \hfill
    \subfigure[The accuracy (\%) with different thresholds on AffectNet7 and AffectNet8.]{
        \includegraphics[scale=0.45]{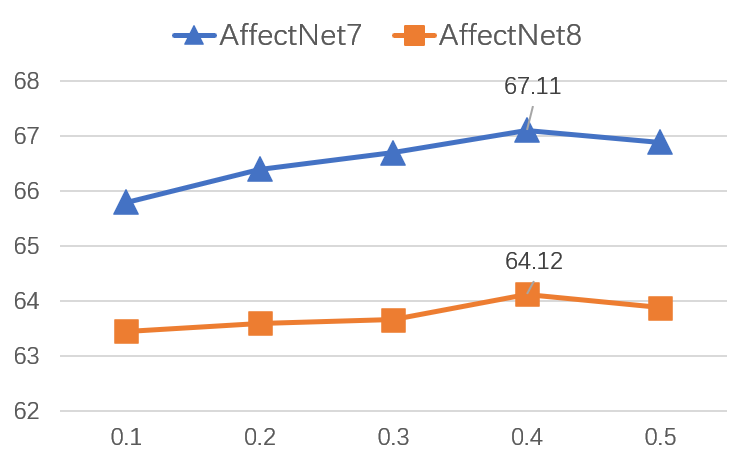}
        \label{threshold_a}
    }
    \caption{Accuracy with different thresholds on multiple datasets.}
    \label{fig:thresholds_all}
\end{figure}

\begin{figure}[!htb]    
\centering
\includegraphics[scale=0.35]{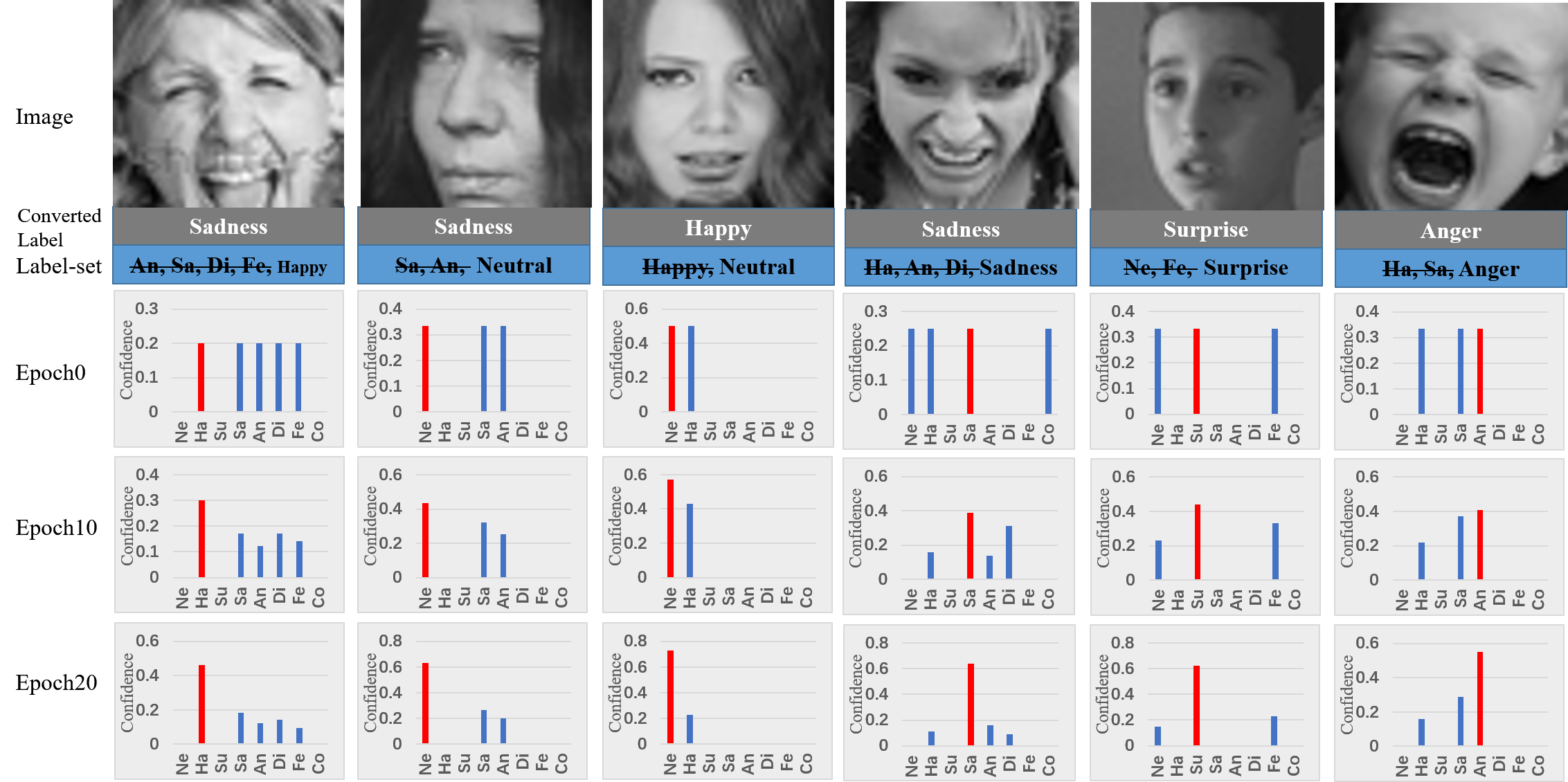}
\caption{The grey part of the figure shows the converted label, equivalent to GT, which is selected by the annotators from the crowdsourcing results in FERPlus.
The blue part shows the candidate set we constructed for the samples corresponding to the converted (GT) labels, where the strikethroughs mark the results of PLL disambiguation for error removal. The predicted labels are consistent with our intuition.}\label{ferplusresult}
\end{figure}

\begin{figure*}[!htb]    
\centering
\includegraphics[scale=0.35]{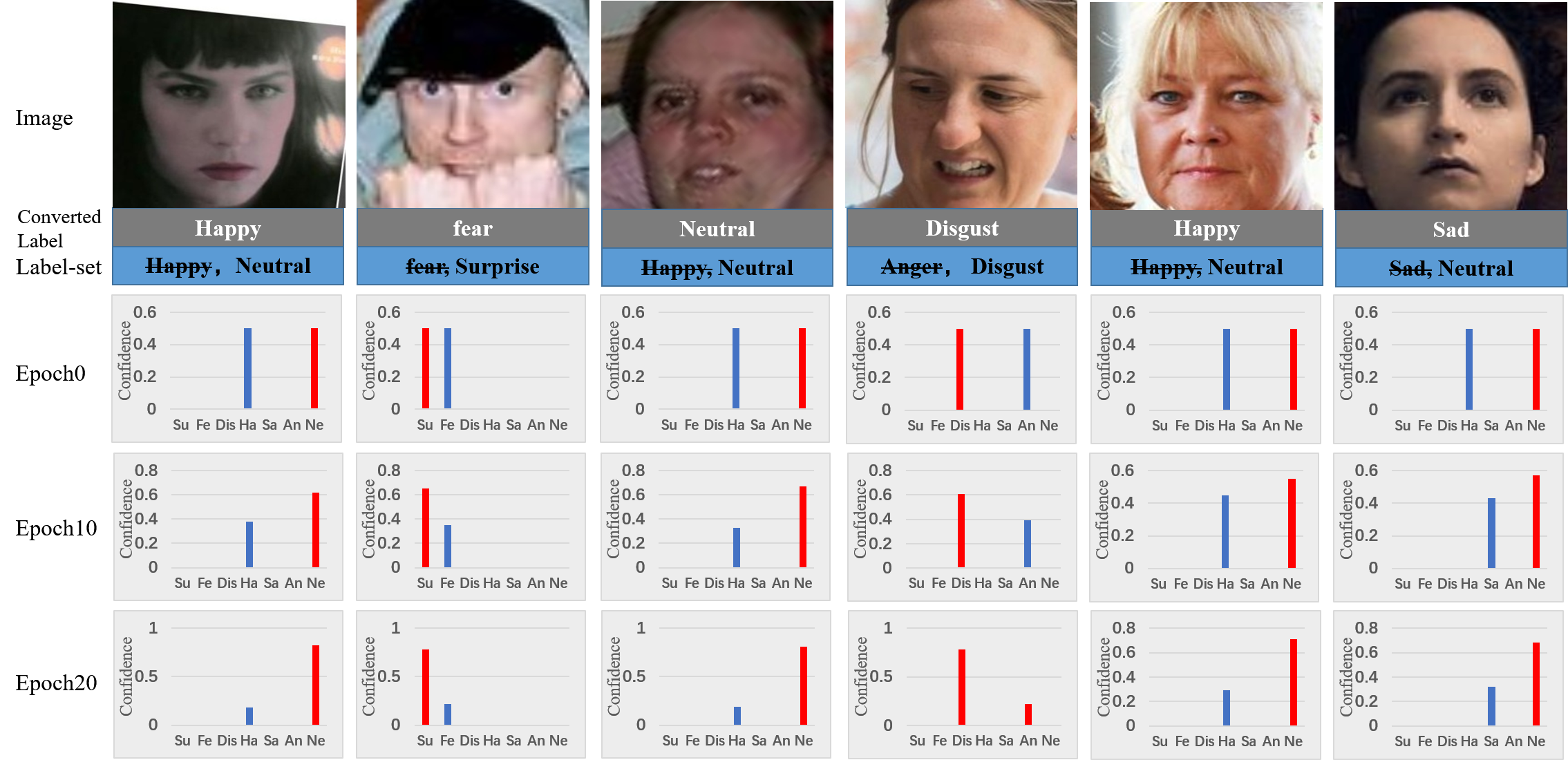}
\caption{The grey part of the figure shows the converted label, equivalent to GT, which is selected by the annotators from the crowdsourcing results in FERPlus. The blue part shows the candidate set we constructed for the samples corresponding to the converted(GT) labels, where the strikethroughs mark the results of PLL disambiguation for error removal. The predicted labels are consistent with our intuition.}\label{sup_raf}
\end{figure*}

\begin{figure*}[t]    
\centering
\includegraphics[scale=0.35]{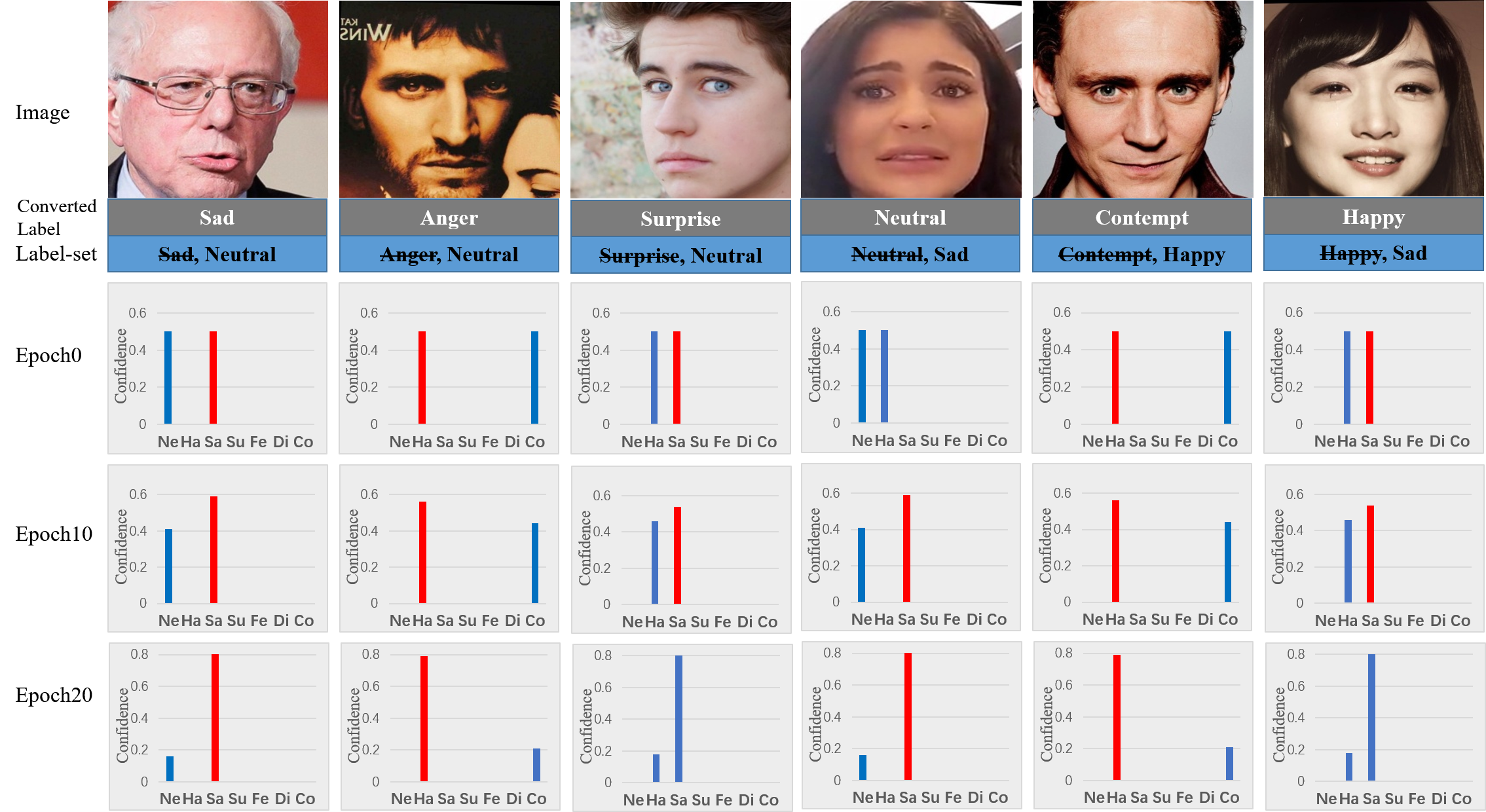}
\caption{The grey part of the figure shows the converted label, equivalent to GT, which is selected by the annotators from the crowdsourcing results in FERPlus. The blue part shows the candidate set we constructed for the samples corresponding to the converted(GT) labels, where the strikethroughs mark the results of PLL disambiguation for error removal. The predicted labels are consistent with our intuition.}\label{sup_a7}
\end{figure*}

The revision confidence module relies on the output of the transformer decoder. In addition, the disambiguation ability of PLL is determined by setting a threshold for the uncertainty of the candidate set. We aim to find and revise the sample, whose confidence is less than the threshold. To evaluate the influence of different thresholds for disambiguation. We conduct an ablation study as \cref{threshold_rf} and \cref{threshold_a} show.

\subsubsection{Ablation for Temperature Hyperparameter $\tau$ and Calculation of the Confusion Matrix}
\label{exp:aba:tau}
To correct the effects of unbalanced data classes, we propose $\mathcal{L}_{uniform}$, whose aim is to distribute the query embedding uniformly over the hypersphere to help the decoder better find the correlation between features and category labels. Inspired by \cite{wang2021understanding}, a small temperature coefficient is beneficial to distinguish the embedding of one category from the embedding of other categories. To verify its validity, we conducted an ablation study, as \cref{temperature} shows.

\begin{table}[htbp]
  \centering
  \caption{The accuracy (\%) with different $\tau > 0$ on RAF-DB and FERPlus.}
  \setlength{\tabcolsep}{4mm}{
    \begin{tabular}{ccccc}
    \toprule
    $\tau > 0$ & 0.001 & 0.005 & 0.01 \\
    \midrule
    RAF-DB  & \textbf{92.35}  & 91.78  & 91.52 \\
    FERPlus  & \textbf{91.41}  & 90.67  & 90.18 \\
    \bottomrule
    \end{tabular}}
  \label{temperature}
\end{table}

\subsection{Influence of parameter K}\label{app:K}
We conducted the influence of K in different settings.
It is somewhat sensitive, and we have set k=2 for optimal performance.
After conducting additional experiments, we observed that in 95\% of cases, a true label is included for k=2, and this increases to 99\% for k=3. In terms of performance, on the RAF-DB dataset, the recognition rates are 92.05\% for k=2, 91.94\% for k=3, and 90.01\% for k=4. It is evident that k is a bit sensitive to different settings, which is possibly due to the trade-off between increasing the possibility of including potential ground truth labels and the simultaneous increase in uncertainty and noise. Consequently, as k increases, the performance gradually declines.

\begin{figure*}[t]    
\centering
\includegraphics[scale=0.45]{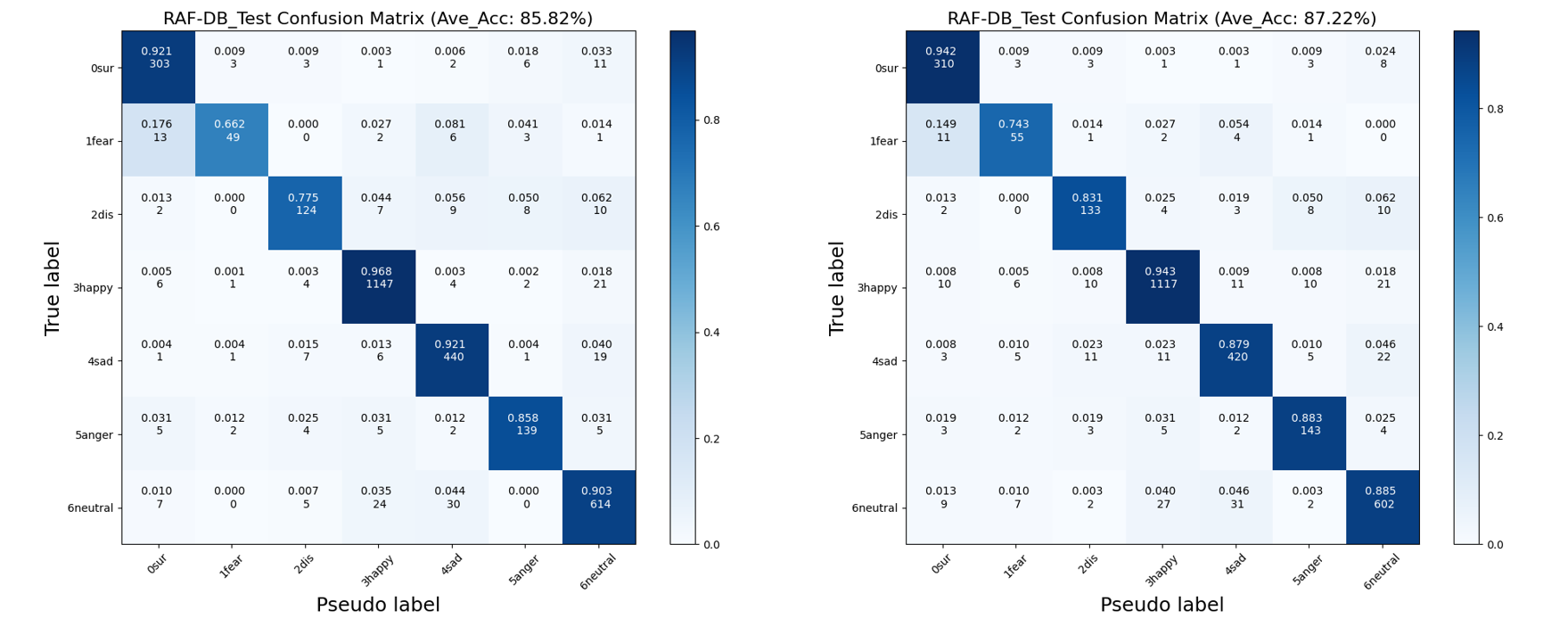}
\vspace{-0.4cm}
\caption{The confusion matrix of the RAF-DB. Left part \textbf{without} $\mathcal{L}_{uniform}$, right part \textbf{with}  $\mathcal{L}_{uniform}$.}\label{tmetrixR}
\vspace{-0.3cm}
\end{figure*}

\begin{figure*}[t]   
\centering
\includegraphics[scale=0.45]{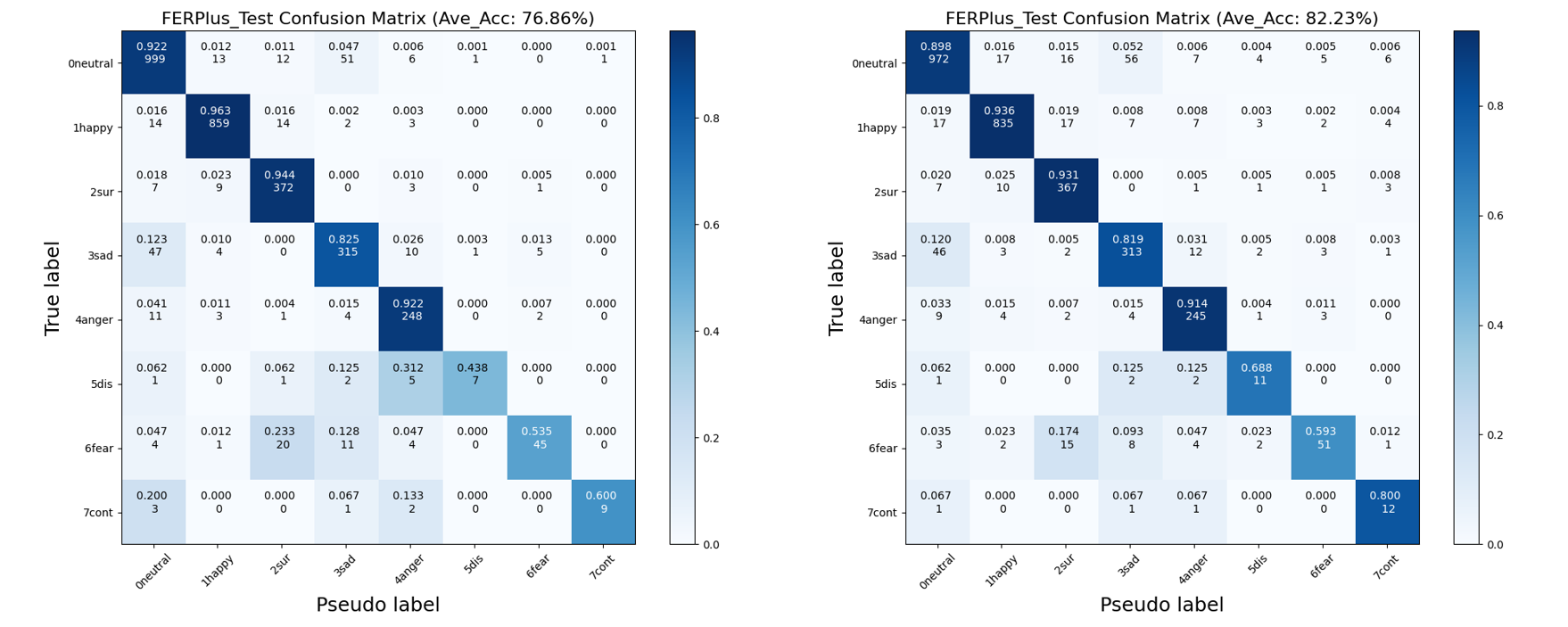}
\vspace{-0.4cm}
\caption{The confusion matrix of the FERPlus. Left part \textbf{without} $\mathcal{L}_{uniform}$, right part \textbf{with}  $\mathcal{L}_{uniform}$. }\label{tmetrixF}
\end{figure*}

To demonstrate the validity of $\mathcal{L}_{uniform}$ more intuitively, we calculated the confusion matrix for each category and calculated overall
We show the results for the RAF-DB and FERPlus confusion matrix with respect to $\mathcal{L}_{uniform}$ in Figure \ref{tmetrixR}, Figure \ref{tmetrixF}, respectively. 

From the experimental results, the temperature hyperparameter $\tau$ reaches the best performance at equal to 0.001, which demonstrates smaller temperature coefficients lead to a more uniform embedding space for query embedding, specifically in that the different classes of query embeddings on the hypersphere move away from each other.

From the confusion matrix on RAF-DB and FERPlus, we improve average accuracy on RAF-DB and FERPlus by 1.4\%, 5.37\%. In the head class, such as the "HAPPY" in RAF-DB, more test samples are classified to the tail class after using the $\mathcal{L}_{uniform}$ constraint, which is beneficial to side effects of class imbalance and helps to eliminate bias to head class in the PLL disambiguation process; for FERPlus, the average accuracy improvement is relatively large, due to the fact that its test samples are in the tail class, such as contempt, with only 15 images, which is an order of magnitude difference compared to the head class. When calculating the average accuracy, the improvement in accuracy in the tail class contributes significantly to the overall average precision. 

\section{Conclusion}
In this paper, we rethink the existing training paradigm and propose that it is better to use weakly supervised strategies to train FER models with original ambiguous annotation. 
To solve the subjective crowdsourcing annotation and the inherent inter-class similarity of facial expressions, we model FER as a partial label learning (PLL) problem, which allows each training example to be labeled with an ambiguous candidate set. 
Specifically, we use the Masked Image Modeling (MIM) strategy to learn the feature representation of facial expressions in a self-supervised manner, even in the presence of ambiguous annotations. 
To bridge the gap between the enriched feature representation and the partial labels, we introduce an adaptive transformer decoder module. This module is specifically designed to interpret the complex, high-dimensional features extracted by the MIM strategy. 
It dynamically adjusts its focus on the ambiguous candidate label sets, employing a soft attention mechanism to weigh the probabilities of each candidate being the true label. 
Extensive experiments show the effectiveness of our method.
Our extensive experiments, conducted on several benchmark datasets for FER, underscore the superior performance of our proposed method over existing approaches. These results validate our hypothesis that modeling FER as a PLL problem and addressing it with a combination of MIM and transformer-based decoding is a potent strategy for enhancing the performance of FER systems.
In conclusion,
our findings suggest that weakly supervised strategies, augmented by self-supervised feature learning and transformer-based decoding, hold great promise for advancing the state-of-the-art in facial expression recognition. 

\section{Safe and Responsible Innovation Statement}
This work aims to improve the accuracy and robustness of facial expression recognition systems. We use ethically sourced and consented datasets, ensuring no sensitive or personally identifiable information is involved. To mitigate potential biases, we consider demographic diversity during model training and evaluation. As facial expression recognition can be misused for surveillance or behavioral manipulation, we emphasize that its deployment must comply with ethical guidelines and legal standards, promoting human-centered and responsible applications.





\bibliographystyle{ACM-Reference-Format}
\bibliography{sample-base}

\appendix

\end{document}